\documentclass[
]{ceurart}

\usepackage{listings}
\begin{document}

\copyrightyear{2022}
\copyrightclause{Copyright for this paper by its authors.
  Use permitted under Creative Commons License Attribution 4.0
  International (CC BY 4.0).}

\conference{D2R2'22: International Workshop on Data-driven Resilience Research 2022, July 07, 2022, Leipzig}

\title{The Ethical Risks of Analyzing Crisis Events on Social Media with Machine Learning}

\author[1,2]{Angelie Kraft}[%
orcid=0000-0002-2980-952X,
email=angelie.kraft@uni-hamburg.de,
url=https://krangelie.github.io/,
]
\cormark[1]
\address[1]{Universität Hamburg, Department of Informatics,
  Vogt-Kölln-Straße 30, 22527 Hamburg}
\address[2]{Hamburger Informatik Technologie-Center e.V. (HITeC),
  Vogt-Kölln-Straße 30, 22527 Hamburg}

\author[1,2]{Ricardo Usbeck}[%
orcid=0000-0002-0191-7211,
email=ricardo.usbeck@uni-hamburg.de,
url=https://www.inf.uni-hamburg.de/en/inst/ab/sems/people/ricardo-usbeck.html,
]
\cortext[1]{Corresponding author.}

\begin{abstract}
    Social media platforms provide a continuous stream of real-time news regarding crisis events on a global scale. Several machine learning methods utilize the crowd-sourced data for the automated detection of crises and the characterization of their precursors and aftermaths.
    Early detection and localization of crisis-related events can help save lives and economies. Yet, the applied automation methods introduce ethical risks worthy of investigation --- especially given their high-stakes societal context. This work identifies and critically examines ethical risk factors of social media analyses of crisis events focusing on machine learning methods. We aim to sensitize researchers and practitioners to the ethical pitfalls and promote fairer and more reliable designs.
\end{abstract}


\maketitle

\section{Introduction}

    Social media platforms are a bottom-up community-driven means for real-time information exchange during crisis events~\cite{palen_online_2008}. They are an important tool in keeping citizens and authorities up-to-date in urgent situations~\cite{anbalagan_chennaifloods_2016,reuter_fifteen_2018}. The shared information can help to establish precautionary measures, organize humanitarian aid, or keep track of missing people.
    Algorithmic approaches are used to efficiently filter, condense, and extract large amounts of social media posts~\cite{dwarakanath_automated_2021,ogie_artificial_2018}. Respective systems nowadays largely rely on deep learning (DL) methods for natural language processing (NLP)~\cite{liu_crisisbert_2021}, computer vision (CV)~\cite{dewan_towards_2017}, or multimodal techniques~\cite{imran_using_2020}. 
    
    The COVID-19 pandemic is a contemporary example where privacy and personal liberties were sacrificed for the quick development of new technologies~\cite{tzachor_artificial_2020}. Although crisis events ask for fast responses, the innovation process must not happen at the cost of ethical considerations.  In this paper, we identify the main ethical risks when analyzing social media content via machine learning (ML) to detect and characterize crises. To scrutinize ethical aspects of technology, we take on a sociotechnical view~\cite{selbst_fairness_2019}: We consider algorithms, their in-, and output data, as well as the social system within which these are embedded.  At the heart of this assessment is the potential long-term impact on people's well-being, values, expectations, and fair treatment, and ultimately on whom a computer system serves and whom it harms.
    We elaborate on each of the risks to sensitize practitioners and researchers developing and deploying respective systems.

\section{Related Work}
\label{rel_work}
    For several years now, ML methods have been used for the analysis of social media posts regarding various types of natural disasters, like floods, hurricanes, earthquakes, fires, and draughts around the globe~\cite{ogie_artificial_2018}. Systems have been developed to facilitate early warnings and to support disaster responses or damage assessments~\cite{dwarakanath_automated_2021}.
    NLP methods can help to distinguish informative from uninformative texts posted on social media, classify the type of crisis event the text belongs to~\cite{liu_crisisbert_2021,DBLP:journals/access/ZaheraJSN21}, or the type of crisis-related content that is discussed (e.g., warnings, utilities, needs, affected people~\cite{dwarakanath_automated_2021}). The same can be done based on photos through CV approaches~\cite{imran_using_2020}. 
    The semantic content of posts can be further leveraged with spatial and/or temporal information to facilitate crisis mapping. For the Chennai flood in 2015, Anbalagan and Valliyammai~\cite{anbalagan_chennaifloods_2016} built a crisis mapping system that classified related tweets regarding their content type (e.g., requests for help, sympathy, warnings, weather information, infrastructure damages, etc.). This information was combined with the geographic coordinates derived from textually mentioned locations via geoparsing. Tools like this which can identify and locate a crisis-related event can help emergency responders navigate complex information streams.
    
    In 2015, Crawford and Finn~\cite{crawford_limits_2015} outlined different classes of limitations of using social media data in crisis informatics. \textbf{Ontological limitations}: Social media activities spike around more sensational instances, although crises onsets are oftentimes followed-up by long-term effects. So, the time frame of a virtual \textbf{discourse is not representative} of the actual crisis timeline.  Further, applications for humanitarian aid have in the past demonstrated a risk of reifying \textbf{power imbalances}: ``Although crowdsourcing projects can allow the voices of those closest to a disaster to be heard, some projects most strongly enhance the agency of international humanitarians'' (p. 495,~\cite{crawford_limits_2015}). \textbf{Epistemological limitations}: The interpretability of social media data is limited by the role that platforms play in shaping the data. Recommendation systems determine what users get to see and share. Moreover, a platform can be seen as a cultural context, with its trends and communicative patterns. Contents may exaggerate real events and be charged with opinion and emotion. Finally, distinguishing between human- and bot-generated messages is not always feasible. \textbf{Ethical issues}: The main point here is the issue of \textbf{privacy}. Personal statements of users are gathered at a time in which they are especially vulnerable. Their posts oftentimes include sensitive information about location or well-being and the needs of themselves or others. Crawford and Finn~\cite{crawford_limits_2015} claim that consent must not be sacrificed for ``the greater good''. 
    
    The privacy issue was also listed as one ethical risk factor by Alexander~\cite{alexander_social_2014}, alongside the loss of discretion caused by a tendency for sharing intimate details. Moreover, the author pointed out that especially wealthy and technologically literate individuals benefit from digital means of disaster management. This adds to the previously mentioned reification of power imbalances. Finally, the spread of rumors and misinformation through users, as well as ideology-driven governance of platforms affect the reliability of details and can cause an overall \textbf{misrepresentation of crises} and their causes.  
    
    Regarding the use of artificial intelligence (AI) in crisis informatics, Tzachor et al.~\cite{tzachor_artificial_2020} highlight issues of the \textbf{disparate impact} of algorithmic outputs, as well as  the lack of \textbf{transparency} and \textbf{trustworthiness} of AI models. The authors demand a principle of \textbf{ethics with urgency}~\cite{tzachor_artificial_2020} which entails (1) \textbf{``ethics by design''} to consider ethical risks throughout the development process and foresee broader societal impacts, (2) validated \textbf{robustness of AI systems}, and (3) building \textbf{public trust} through independent oversight and transparency.

\section{Ethical Risks}

    The presented work consolidates previous ethical risk assessments of crisis informatics with social media data (Section \ref{rel_work}) with an emphasis on ML methods. We expand on previous works by examining recent technological advancements and newer insights on their potential risks. For a better overview, the following sections are sorted by data- and algorithm-related concerns. Please note that there is a conceptual overlap between some of the issues mentioned: e.g., limited representativeness of data is problematic because algorithms capture and reproduce biases~\cite{bender_dangers_2021}. However, awareness of the problem layers allows for an in-depth understanding and faceted scrutiny of future software.

\subsection{Limited Representativeness}
\label{data}
    To understand who communicates and receives information on social media, it is necessary to take a disaggregated look at user demographics. 
    In 2020, there were more than 3.6 billion social media users worldwide.\footnote{\scriptsize\url{https://www.statista.com/statistics/278414/number-of-worldwide-social-network-users/}} Facebook ranks first amongst the most popular platforms, with 2.9 billion users as of January 2022.\footnote{\scriptsize\url{https://www.statista.com/statistics/272014/global-social-networks-ranked-by-number-of-users/}}  Even though Twitter did not make the top ten list with only 426 million users, it is still the most researched social media platform~\cite{dwarakanath_automated_2021}. The reason for this might be its easily accessible API for researchers, allowing them to analyze its full stream of posts. By far margin, the majority of Twitter users come from the United States or Japan (India ranked third with less than half of the amount of users in Japan, as of January 2022).\footnote{\scriptsize\url{https://www.statista.com/statistics/242606/number-of-active-twitter-users-in-selected-countries/}} In April 2021, 38.5\% of all Twitter users ranged between ages 25 and 34, and 21\% were between 35 and 49 years old.\footnote{\scriptsize\url{https://www.statista.com/statistics/283119/age-distribution-of-global-twitter-users}} 
    These numbers indicate that most research done on Twitter corpora is based on the \textbf{perceptions of a non-representative sample of people}. Here, perception relates to both the reality witnessed by individuals due to spatio-temporal factors, and also to belief and ideology -- especially in the context of crisis~\cite{landau2004deliver}. 
    
    Social media platforms use recommendation systems to display content that echoes users' interests and opinions. The \textbf{filter bubble hypothesis} states that this mechanism leads to isolated \textbf{echo chambers} and polarization of social networks~\cite{difranzo_filter_2017}.
    Regarding the attention dynamics on social media, some voices recently argued that the Twitter community paid more attention to the 2022 Ukraine crisis than other wars and genocides happening in the meantime.\footnote{\scriptsize\url{https://www.npr.org/sections/goatsandsoda/2022/03/04/1084230259}} They claim that such phenomena stem from and reinforce global power inequalities. Social media attention propagates to mainstream media and governments, and affects decisions regarding humanitarian aid.
    
    Following the principle of equity, \textbf{we suggest an over-emphasis on minority and disadvantaged groups} during software development, instead of targeting a representative sample. After all, we should focus on those who rely most on humanitarian aid and crisis relief.

\subsection{Misinformation}
\label{rumors}
    The impact that misinformation can have on societies became evident in the COVID-19 pandemic. As mentioned in~\cite{tasnim_impact_2020}, false social media rumors about a lockdown in the United States inviting civilians to stockpile certain products -- a behavior that affects supply chains and causes demand-supply gaps~\cite{sukhwani2020covid}. While households with higher socioeconomic status are more able and likely to stockpile~\cite{o2021preparing}, low-income households are prone to food insecurity due to decreased availability and increased prices~\cite{dasgupta2022impact}. 
    
    Misinformation comprises different forms of \textbf{intentionally and unintentionally false or inaccurate information}~\cite{wu_misinformation_2019}: e.g., disinformation (intentional), rumors, fake news, urban legends, spamming and trolling. Through an analysis of ca. 126,000 news items shared via Twitter between 2006 and 2017, Vosoughi et al.~\cite{vosoughi_spread_2018} found false rumors transmitted ``farther, faster, deeper, and more broadly than the truth in all categories of information'' (p. 2). During the COVID-19 pandemic, inaccurate  social media posts about infection prevention were shared more often than accurate posts~\cite{obiala2021covid}. Thus, misinformation can be an obstacle to establishing containment measures~\cite{tasnim_impact_2020}. Moreover, it can unnecessarily trigger public fear.
    In the 2014 Ebola epidemic,  a majority of the misinforming tweets exaggerated the spread and fatality of the disease~\cite{sell_misinformation_2020}.

    Automated analysis systems must incorporate a mechanism that discards false information to \textbf{avoid its consolidation and dissemination}, as well as the conjuring of inappropriate coping measures. While the identification of misinformation is mainly done through expert coding, ML solutions are on the rise~\cite{hunt_monitoring_2020}. Detecting misinformation directly through visual or textual content is difficult.  This is why some works also incorporate contextual information, like temporal patterns of posting (posts published in bursts), propagation through social networks, or hashtags~\cite{wu_misinformation_2019}.

\subsection{Privacy}
    The availability of personal information on the web does not obviate the \textbf{need for unambiguous consent} regarding its collection and use through a third party~\cite{zimmer_but_2010}. Informed consent about third-party use is usually provided by the user as a prerequisite to using the platform. But even then, users might not be fully aware of what their consent entails~\cite{hemphill_saving_2021}. Furthermore, the willingness to consent is subject to change and, according to GDPR law, the agreement is retractable by the user. Hence, published corpora -- such as   CrisisLex~\cite{olteanu_crisislex_2014} -- must be taken down or altered if users wish to remove their data. Yet, that is in practice hardly feasible: ML training corpora are downloaded, copied, and shared, ML models are trained on them, and certain posts might be cited in publications. 
    Direct quotes in public datasets may be traceable and allow identification of the author's identity.\footnote{\scriptsize\url{https://www.ucl.ac.uk/data-protection/sites/data-protection/files/using-twitter-research-v1.0.pdf}} The Chennai flood crisis mapping system~\cite{anbalagan_chennaifloods_2016} mentioned earlier geoparsed user posts to derive geographic coordinates from textual location mentions. Even though these were at the time posted to alert readers regarding events happening at certain places, it is questionable whether consent was provided for the type of post-processing done by the authors. 
    
    \textbf{During times of crisis, the data shared publicly on social media are especially personal}. The very content of crisis-related information spans from the date and location details, characterizations of individuals including names and imagery and reports about physical and psychological harms to utterances of grief and fear~\cite{alexander_social_2014}. At trying times, people are at their most vulnerable. Collecting their crisis-related posts without dedicated consent and careful consideration, thus, severely violates individual privacy~\cite{crawford_limits_2015,tzachor_artificial_2020}. 

\subsection{Algorithmic Bias}
    Crises affect vulnerable groups disproportionately and biased automation systems risk exacerbating this dynamic. In the recent COVID-19 health crisis, for instance, the hasty development and non-peer-reviewed publishing of socially \textbf{biased algorithms have contributed to social inequalities} between black and white communities in the United States~\cite{roosli_bias_2021}.  
    
    In the context of social media, sentiment classifiers have been used to determine the emotional state of the public during crises, estimate overall social impact, or filter individual posts for urgency~\cite{kaur_sentiment_2015,zhang_semiautomated_2020}. However, available sentiment analysis systems are to a large extent socially biased~\cite{kiritchenko_examining_2018}. It was pointed out by Yang et al.~\cite{yang_towards_2020} that this is one of the major bias-introducing factors in disaster informatics. Not only content but also dialect can yield bias~\cite{blodgett_racial_2017}: for example, hate speech detectors may overestimate the toxicity of sentences in African-American Vernacular English~\cite{sap_risk_2019}. Most large-scale language models like BERT~\cite{devlin2018bert}, GPT~\cite{gpt}, and their successors have been shown to reproduce a variety of biases and stereotypes~\cite{bender_dangers_2021}. Their extensive use across NLP tasks makes \textbf{social bias a general issue in this domain}. 
    
    Another source of bias identified by Yang et al.~\cite{yang_towards_2020} is yielded by the positive correlation between socioeconomic status and tweet density~\cite{li2013spatial}. Some crisis event detection systems consider temporal patterns, like bursts of tweets. Such systems might detect events disproportionately more in wealthier areas~\cite{yang_towards_2020}. We earlier mentioned the privacy risk of geoparsing. Besides that, another ethical limitation arises from biases in the technologies used in geoparsing pipelines~\cite{yang_towards_2020}. Named-entity recognition (NER) systems detect  names of places, subjects, and the like in texts. These were found to exhibit binary gender bias, i.a. the accuracy with which female entities are identified was lowered. Yang et al.~\cite{yang_towards_2020} suspect that NER systems might also be prone to other types of sociodemographic biases. Finally, the authors claim that CV systems for disaster characterization, like flood depth estimators~\cite{bai2016weibo}, might yield disparate outcomes. That is because some of these systems use humans for scale reference. However, models for the recognition of human features output differently accurate results for different social groups~\cite{buolamwini2018gender,du2021fairness}.  To avoid the disparate impact of biased algorithms, ML systems must undergo in-depth auditing, for example,via disaggregated performance evaluation~\cite{buolamwini2018gender}.

\subsection{Availability of Machine Learning Technologies}

    Modern ML technologies mostly serve developed countries~\cite{dwarakanath_automated_2021}, both in terms of availability and fit. As mentioned earlier, algorithms are often biased or inaccurate for whole demographic groups \cite{bender_dangers_2021}. This in effect not only derogates people but also prevents them from benefiting from technological progress.  
    
    During our research, we noticed that non-English crisis corpora are not easily found \cite{DBLP:journals/tweb/RudraGGG18}. The same counts for well-performing language models and other NLP applications. The \textbf{neglect of so-called ``low-resource'' languages} (language for which digital textual content is less available or has not been systematically gathered) is a widely discussed issue. 88\% of all languages are completely ignored by NLP research, with no hope for change anytime soon~\cite{joshi2021state} despite a growing amount of (European) platforms for NLP systems~\cite{DBLP:conf/lrec/RehmGLPWUKDGFCF20}. Those who suffer most from crises and would particularly benefit from support and prevention systems are least likely to be considered during development. With this, social inequalities are further reified.

\subsection{Lack of Transparency}
    We have discussed different factors affecting the reliability of ML systems:  unrepresentative and nonfactual data  (from intentional and unintentional misinformation), algorithmic bias, and a lack of fit to most language or cultural regions. To complicate matters further, most ML and AI applications are \textbf{non-transparent decision makers}. So, irregularities are not easily spotted by non-technical personnel. The black-box nature of these complex models is a restricting factor, especially in a high-stakes crisis. 
    
    To improve the transparency of research and development, open-source and open-data practices have emerged. Public availability of training data facilitates scrutiny of a model's potential limitations. However, this habit conflicts with the fact that social media data created during emergencies are particularly sensitive (see above). While reproducibility is certainly an important control mechanism, strict guidelines and compliance of practitioners are needed to ensure that heightened privacy needs are met. 
    
    Finally, we emphasize the need for explainable and interpretable ML methods. The inability to trace why a system suggests certain decisions \textbf{limits public control and legitimization}~\cite{tzachor_artificial_2020}. Authorities and civil persons should be able to comprehend the reasons behind algorithmic decisions -- to act justifiably and not fall victim to an algorithmic fallacy. Hybrid AI methods - combining DL and Knowledge Graphs~\cite{DBLP:journals/csur/HoganBCdMGKGNNN21} - require less data, are explainable through their ground, and can therefore be used more effectively and efficiently in sensitive areas~\cite{DBLP:journals/apin/EbrahimiEBH21}. 

\section{Future Directions}

    As claimed in~\cite{tzachor_artificial_2020}, ML systems for crisis need ``ethics with urgency'': (1) Ethical issues must be considered from the outset by foreseeing the system's societal impact, (2) systems must be robust, and (3) 
    public trust through independent oversight and transparency must be built. We suggest evaluating crisis technology as sociotechnical systems~\cite{selbst_fairness_2019}: algorithms are embedded in social and political dynamics which pose ethical requirements to the data and algorithmic outcomes. Understanding the stakeholders' values and needs, the long-term effects of the system on society (and vice versa), and context-specific societal demands during a crisis becomes an essential element of software development. 
    
    Developers and researchers should consider where and how the data were collected and whose experiences and motivations they represent. ``Datasheets for datasets'' \cite{gebru2021datasheets} can help guide in-depth examination of the data and shed light on potential risk factors. The resulting datasheet is an accompanying artifact to the developed system allowing for transparency and accountability later on. 
    Similarly, we recommend the use of model cards~\cite{mitchell2019modelcards} to transparently document how and on which data an ML model was developed and evaluated, what its technical and ethical limitations are, as well as its intended use. To circumvent disparate impact, data and algorithms should undergo bias auditing, for example through disaggregated performance analyses and the use of suitable fairness metrics. The choice of fairness metric again heavily relies on the social setting of the system~\cite{selbst_fairness_2019}. We suggest putting disadvantaged groups at the focus of crisis technology to approach equity and help those particularly affected by crises. Moreover, we encourage examining whether or not a planned technical solution is appropriate in the given situation, to begin with, and avoid technosolutionism~\cite{selbst_fairness_2019}. If ethical risks are inevitable, abandoning an idea must be considered as an option. All in all, given its context-specific nature, there is no standard solution for ethically developing crisis technology. This must be judged on a case-by-case basis.

\section{Conclusion}

    ML-based analysis of social media streams can facilitate a swift aggregation and filtering of information during crises. This can support civilians, emergency responders, and authorities to cope quickly. However, the pairing of social media-sourced data and ML algorithms gives rise to several ethical risks. In this position paper, we addressed issues of representativeness, factuality, and privacy of social media data, ML algorithms' proneness to reproduce bias, as well as their unavailability for many languages and cultures, and their lack of transparency. Furthermore, the vulnerability of social media users during crises is increased. This results in heightened ethical requirements for crisis informatics systems to secure people's well-being. We conclude that the harms otherwise disproportionately affect already disadvantaged groups. Future work must focus on supporting these very groups to strive for equity. 
    
     To be able to fulfill the inherent goal of helping those in need, practitioners must examine all facets of the impact their software is going to have in the long run. Rapid development at the cost of ethics will else paradoxically defeat its purpose.  

\begin{acknowledgments}
    The authors acknowledge the financial support by the Federal Ministry for Economic Affairs and Energy of Germany in the project CoyPu (project number 01MK21007[G]).
\end{acknowledgments}

\bibliography{bib}

\end{document}